\documentclass[letterpaper, 10 pt, conference]{ieeeconf}  

\IEEEoverridecommandlockouts                              

\usepackage{tabularx}
\usepackage{multirow}
\usepackage{graphicx}
\usepackage{xcolor}
\usepackage{subcaption}
\usepackage{amsmath} 
\usepackage{subcaption}
\newcommand{\ourmethod}{\emph{PaceForecaster}}

\title{\LARGE \bf
Fast Navigation Through Occluded Spaces via Language-Conditioned Map Prediction
}

\author{Rahul Moorthy Mahesh, Oguzhan Goktug Poyrazoglu, Yukang Cao, Volkan Isler%
}

\begin{document}

\maketitle
\thispagestyle{empty}
\pagestyle{empty}

\begin{abstract}
In cluttered environments, motion planners often face a trade-off between safety and speed due to uncertainty caused by occlusions and limited sensor range.
In this work, we investigate whether co-pilot instructions can help robots plan more decisively while remaining safe. We introduce \ourmethod{}, as an approach that incorporates such co-pilot instructions into local planners. \ourmethod{} takes the robot’s local sensor footprint (Level-1) and the provided co-pilot instructions as input and predicts (i)~a forecasted map with all regions visible from Level-1 (Level-2) and (ii)~an instruction-conditioned subgoal within Level-2. The subgoal provides the planner with explicit guidance to exploit the forecasted environment in a goal-directed manner. We integrate \ourmethod{} with a Log-MPPI controller and demonstrate that using language-conditioned forecasts and goals improves navigation performance by 36\% over a local-map-only baseline while in polygonal environments.
\end{abstract}

\section{Introduction}
\begin{figure}[h!]
  \centering
  \includegraphics[trim=4.5cm 6.0cm 9.0cm 2.0cm, clip, width=\linewidth]{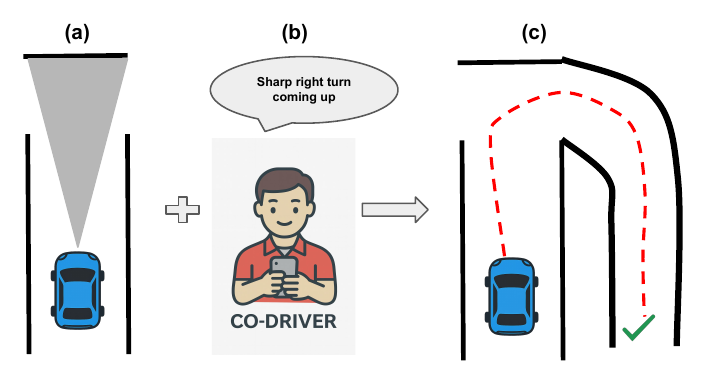}
  \caption{Co-Driver Setting: (a) Assume the robot approaches a junction with occlusions in both left and right with no information on the direction and the sharpness of the turn. (b) The co-driver gives the instruction of ``Sharp Right Turn". We seek to answer the question - (c) Can we generate the turn and the goal (shown in green tick mark) based on the language instruction of the co-driver so that the car can generate the path shown in red.}
  \label{fig:placeholder}
\end{figure}
Robot navigation in cluttered environments in a fast \emph{and} safe manner is challenging due to  uncertainties induced by the environment as well as the robot's controls. In some settings, the environment can be mapped in advance, and fast localization methods can be employed onboard. In such settings, either traditional trajectory optimization methods~\cite{ratliff2009chomp}, kinodynamic planners~\cite{lavalle2001randomized} or more recent Model Predictive Control (MPC) methods~\cite{mohamed2022autonomous} can be employed to obtain optimal or near-optimal trajectories. If a map is not available, Simultaneous Localization and Mapping (SLAM) algorithms~\cite{labbe2019rtab} can be employed to obtain a map of the environment observed up to the current point in time, as well as the robot's pose in this partial map. However, not knowing what is ``around the corner" limits the horizon for trajectory optimization, and therefore induces a safety/speed trade-off. 

In this paper, we consider a novel setting which can be used to mitigate this trade-off using verbal instructions from a co-pilot who can inform the robot about, for example, a ``Sharp right turn coming up." This setting is common in rally racing. More generally, a driver in a new city could be assisted by a local in a passenger seat. In robotics settings, the robot might receive help from a remote supervisor who, rather than teleoperating a single robot, could supervise many robots simultaneously. The main question we seek to answer is how to incorporate this additional high-level information into trajectory planning. 

\begin{figure}[h!]
    \centering
    \begin{subfigure}{0.49\linewidth}
        \centering
        \includegraphics[width=0.49\linewidth]{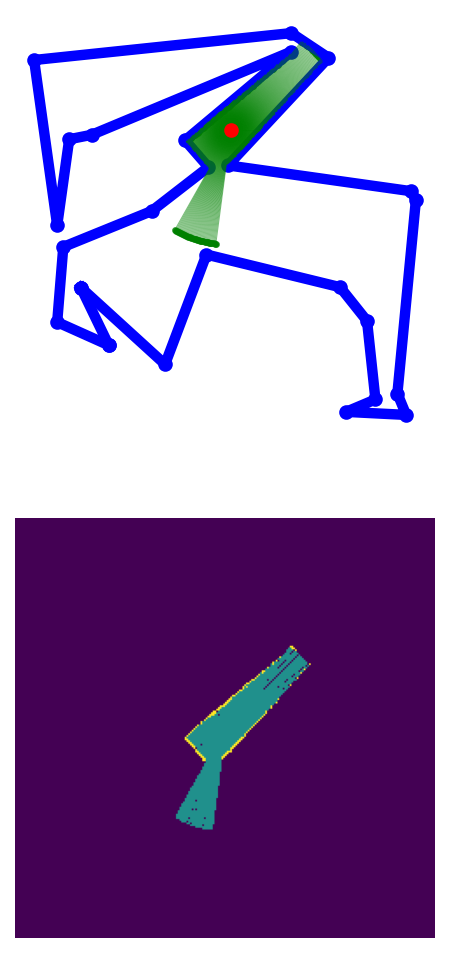} 
        
        \caption{Level-1 Map}
        \label{fig:sub1}
    \end{subfigure}
    \begin{subfigure}{0.49\linewidth}
        \centering
        \includegraphics[ width=0.49\linewidth]{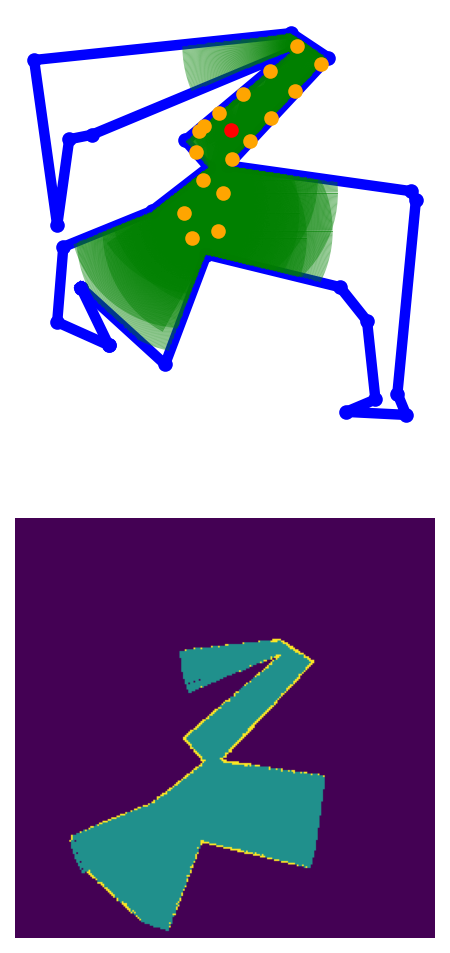}
        \caption{Level-2 Map}
        \label{fig:sub2}
    \end{subfigure}
    \caption{Level-1 vs Level-2 Maps: Level-1 shows the local LiDAR map in the robot frame (robot shown in red). Level-2 is the region that would become visible if a LiDAR scan was taken from all of the frontier endpoints of the Level-1 map (yellow). The bottom figure shows the final occupancy maps of Level-1 and Level-2 where violet represents unknown, blue represents free space and yellow represents obstacle regions.}
    \label{fig:level_1_2}
\end{figure}

Our main contribution is a novel architecture \ourmethod{}, which takes as input the robot's current sensor footprint (which we call the Level-1 region, or L1 for short) and the instructions $I$. Based on this input, \ourmethod{} outputs (i)~what's around the corner: these are points visible from points in L1 (as shown in Figure \ref{fig:level_1_2}). We call this region Level-2, or L2 for short , and (ii)~a subgoal  $G \in L2$ that the robot should navigate toward. Note that the subgoal $G$ is conditioned on the instruction $I$. Therefore, the same input L1 with two different instructions $I_1$ and $I_2$ would output the same L2 map but two different subgoals. Once the new map and subgoal are generated, a model predictive controller is employed to generate the robot's trajectory in real-time. In this paper, we employ Log-MPPI~\cite{mohamed2022autonomous}. We show that predicting L2 and the subgoal $G$ improves the navigation performance of Log-MPPI by \textbf{36\%} compared to using only Level-1 (sensor footprint).
\par In summary, the contributions of our work are:
\begin{itemize}
        \item We propose a novel problem setting where instructions inform local planners about the upcoming turns to improve navigation in complex multi-modal environments.
        \item We present \ourmethod{}, which forecasts regions beyond the sensor horizon by generating Level-2 and predicts the goal with respect to robot in Level-2 conditioned on the instructions.
        \item We provide experimental validation in complex simulated polygon environments and real-world environments, showing that local planners using Level-2 enables safer navigation by \textbf{36\%} at high velocities compared to relying solely on the local map.
    \end{itemize}

\section{Related Work}
We summarize the related work in three areas: Unknown-area exploration, Occlusion Reasoning, and vision-language navigation.
\par \textbf{Unknown-area exploration}: Autonomous exploration focuses on planning a path that maximizes expected information gain. Ho et al.~\cite{ho2025mapexindoorstructureexploration} proposed MapEx, which predicted multiple plausible complete maps from a partial map and used them to select informative exploration frontiers. Gao et al.~\cite{gao2025mappingsenselightweightneural} proposed a lightweight neural network that predicted the unknown portion of the local map and efficiently generated exploration frontiers using the predictions. Zwecher et al~\cite{zwecher2022integratingdeepreinforcementsupervised} proposed use of reinforcement learning to reduce the computational cost of information-gain estimation within an exploration framework. Tao et al.~\cite{tao2023learningexploreindoorenvironments} used predicted maps with a reinforcement learning-based planner to select the best navigation goals that maximizes the expected information gain for safer aerial exploration. Tan et al.~\cite{survey} presented a survey summarizing the field of map-prediction–based exploration. In our work, rather than selecting next-best views to maximize information gain, we study instruction-conditioned navigation: given a verbal instruction, we forecast a Level-2 (L2) occupancy map beyond the field of view (FOV) and generate a robot-frame subgoal that a standard local planner can use under occlusion. 

\par \textbf{Occlusion Reasoning}: Map completion under occlusion is another line of work closely related to ours. Wang et al.~\cite{wang2024agrnav} proposed AGRNav for semantic map completion by capturing contextual
information and occlusion area features in ground-air collaborative setting. Luperto et al.~\cite{luperto2023mapping} proposed exploiting the structure of indoor environments to predict layouts of unobserved rooms behind closed doors. Katyal et al.~\cite{katyal2020highspeedrobotnavigationusing} proposed a method that forecasts occupancy beyond the FOV, enabling high-speed navigation. Sharma et al~\cite{sharma2023proxmapproximaloccupancymap} proposed predicting the occupancy near the vicinity of the robot movement space to enable efficient navigation.
Wei et al.~\cite{wei2021occupancy} proposed using supervision from occupancy grids with a wider FOV to reason about occlusions when the instantaneous FOV was limited. Unlike these methods which rely on availability of environment-specific prior data or wider-FOV supervision. We require no such supervision and infer regions outside the current FOV solely from language instructions.
\par \textbf{Vision-Language Navigation}: The task of following instructions is studied extensively in VLN settings. Huang et al.~\cite{huang2022visual, huang2023audiovisuallanguagemaps} proposed VLMaps and AVLMaps to follow a set of language instructions by grounding them to created vision-language and audio map. Chen et al.~\cite{chen2020topologicalplanningtransformersvisionandlanguage} proposed a method that grounds vision-language instructions in a topological map for navigation. Another line of work in this literature is zero-shot vision-language navigation using pretrained models~\cite{goetting2024end, du2025vl, shah2023lm}, which focused on following language queries grounded by visible landmarks in robots's FOV. Our work differs from them as we reason about the regions outside the current field of view and do not assume the existence of a prebuilt map for grounding.
\section{Problem Formulation}
In this work, we focus on rally racing instructions, in which direction, cumulative turn-angle change (binned into severity levels), and distance to the upcoming turn are represented symbolically. Now, we formulate the problem of following instructions for point-to-point navigation in an unknown environment as follows. At time $t$, let $\mathbf{I_{t}}$ denote the instruction, $\mathbf{M_{t}}$ the Lidar Local Map (Level-1), and $\mathbf{G^{(1)}_{t}}$ as the known subgoal in $\mathbf{M_{t}}$. The objective is to determine language-conditioned subgoal $\mathbf{G^{(2)}_{t}}$ and a control trajectory $T$ to $\mathbf{G^{(2)}_{t}}$ while optimizing for collision-free navigation.
\section{Proposed Approach}
\begin{figure*}[h!]
    
    \centering
    \includegraphics[trim=0.5cm 4.7cm 3.5cm 3.5cm, clip, width=\textwidth]{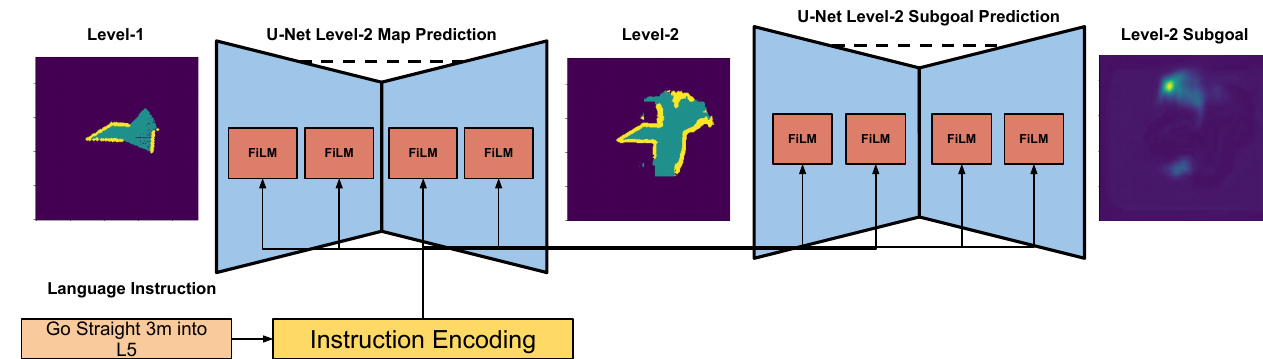} 
    \caption{\ourmethod{} architecture: There are two main modules: \textbf{U-Net Level-2 Prediction} and \textbf{Subgoal Prediction}.
\textbf{U-Net Level-2 Prediction}: First, the Level-1 local map in the robot frame is provided as input to a U-Net, which generates the Level-2 predicted map conditioned on the instruction encoding.
\textbf{Subgoal Prediction}: Given the predicted Level-2 map as input, the subgoal head outputs a heatmap of the subgoal location, also conditioned on the instruction encoding.
Conditioning is applied via FiLM~\cite{perez2017filmvisualreasoninggeneral} to the spatial CNN features in both the map and subgoal prediction networks.
During training, the Level-2 map and subgoal prediction models are optimized jointly.
In the occupancy grid representation of the map, violet denotes unknown regions, yellow denotes obstacles, and blue denotes free space.}
    \label{fig:architecture}
\end{figure*}

In this section, we present the three main modules of \ourmethod{}: Instruction Encoding, Map Prediction, and Subgoal Prediction. Figure~\ref{fig:architecture} shows the end-to-end architecture diagram of the three modules.

\subsection{Instruction Encoding}
We use a one-hot encoding to represent the instructions. Our representation consists mainly of two parts: Direction, and Severity Representation. \textbf{Direction Representation}: We encode each navigation direction as a three-dimensional categorical vector corresponding to straight, left, and right. \textbf{Severity Representation}: Severity represents the turn intensity which is encoded as an integer from 1 to 6, where $6$ denotes a cumulative turn-angle change $<30^\circ$, $1$ denotes $\ge 150^\circ$, and the the remaining severity levels cover successive $30^\circ$ bins. Additionally, we also discretize the distances into three different bins: short ($<5$ m), medium ($<10$ m), and long ($>10$ m). We use a shared six-dimensional one-hot vector to encode both distance bins and turn severity. Lastly, to handle multiple instructions, we concatenate the encodings of individual instructions in sequence to produce a single embedding $\mathbf{E_{I_{t}}}$. The generated instruction embedding is then used to condition both the Level-2 map and the subgoal prediction networks.

\subsection{Level-2 Map Prediction Architecture}
The language instructions contain information about the environment beyond the robot's sensor horizon. Hence, we formulate a Level-2 map prediction task given the LiDAR local map $\mathbf{M_{t}}$ and language encoding $\mathbf{E_{I_{t}}}$. We build on the inpainting architecture presented in~\cite{katyal2020highspeedrobotnavigationusing}. Specifically,  we train a U-Net encoder-decoder architecture to predict Level-2 map $\mathbf{F_{t}}$ given the input $\mathbf{M_{t}}$. We condition the U-Net’s spatial CNN features in both the encoder and the decoder using FiLM~\cite{perez2017filmvisualreasoninggeneral} with $\mathbf{E_{I_{t}}}$. 
\par We represent $\mathbf{M_{t}}$ and $\mathbf{F_{t}}$ as a three-class occupancy grid. Occupied cells are represented as $1$, free cells as $0$, and unknown cells as $-1$. This representation explicitly marks unobserved areas for better supervision. 
Weighted multiclass cross-entropy and dice loss were used as our loss functions to train the prediction architecture. The choice of weighted multiclass cross-entropy is to reduce the influence of unknown class.
We set weights to 5 (occupied) and 2 (free), based on their average ratios to the unknown class. Finally, we apply the classifier-free guidance approach~\cite{ho2022classifier} popularly used in diffusion models for training the inpainting architecture. In particular, we randomly drop instruction encoding conditioning 10\% of the time during training to improve the ability of the model to utilize $\mathbf{E_{I_{t}}}$.  

\subsection{Level-2 Subgoal Prediction Architecture}
Level-2 map prediction alone is insufficient for generating trajectories that leverage look-ahead information because, without an explicit goal, local planners lack the guidance to use the predicted environment. 
We therefore estimate the language-conditioned subgoal $\mathbf{G^{(2)}_{t}}$ in the robot frame. The objective is to predict the target location needed to follow the instruction. For subgoal generation, we use the same architecture as map prediction using $\mathbf{F_{t}}$ as an input. The network predicts the Gaussian goal heatmap centered at the target in the local frame. The model is trained using mean squared error (MSE) loss to encourage the generation of a sharp goal heatmap. The Level-2 subgoal and map prediction model are trained jointly for faster convergence and for making subgoal predictions robust to noise of the map prediction. Finally, we utilize standard local planners to generate $\mathbf{T}$ that executes the language instructions using the predicted map $\mathbf{F_{t}}$, predicted subgoal $\mathbf{G^{(2)}_{t}}$ and Level-1 subgoal $\mathbf{G^{(1)}_{t}}$. 
\par \textbf{Network Architecture and Training Details}: The Level-2 map and subgoal prediction U-Net architecture consists of four encoder and decoder stages. The encoder and decoder channel widths are $\{32, 64, 128, 256\}$ with a $512$-channel bottleneck and skip connections at each scale. We train the models with the Adam optimizer using a learning rate of 1e-4. $\mathbf{M_{t}}$ and $\mathbf{F_{t}}$ are $240\times 240$ cells with resolution of $0.1\,\mathrm{m}$/cell, resulting in a LiDAR range of 12m front and back.
\section{Dataset Generation}
The problem of Level-1 to Level-2 generation requires a dataset of polygons that exhibit diverse visibility scenarios. However, traditional navigation datasets such as BARN~\cite{perille2020benchmarking} and DynaBARN~\cite{nair2022dynabarn} primarily evaluate the ability of local planners to sample diverse trajectories for obstacle avoidance. These datasets define complexity in terms of the number of obstacles rather than the richness of visibility conditions. Similarly, real-world datasets like floorplan~\cite{cruz2021zillow} are restricted to rectilinear polygons, which limit their diversity and generalization in terms of visibility. To address this gap, Moorthy et al.~\cite{moorthy2025visdiffsdfguidedpolygongeneration} recently introduced Visdiff, a new polygon dataset explicitly designed for visibility reasoning that maximizes diversity of visibility graphs by uniformly sampling with respect to link diameter. Hence, in this work, we utilize the Visdiff dataset to generate training pairs of Level-1 and Level-2 maps with associated instructions.

\par The training dataset generation involves subsampling polygons representing 7000 unique visibility graphs from the Visdiff dataset and scaling them to $[-10, 10]^2$. The scaling is performed to simulate a real-world robot in the environment. We then sample two farthest points on the polygon based on maximum link distance and compute a medial-axis path between them to maximize clearance (safety) for navigation. The final training dataset for \ourmethod{} is generated by sampling Level-1 and Level-2 pairs along the medial-axis path. The subgoals and instructions for each sampled pair are generated based on the furthest point of the medial axis path within the Level-2 generated polygon. Additionally, augmentations are performed for subgoal generation task by perturbing the point along the boundary of the Level-2 polygon. The augmentations introduce
the property of multiple subgoals with the same Level-1 and Level-2 pair. The final training dataset consists of 150,000 Level-1 and Level-2 pairs with their respective instructions and subgoals.
\section{Experiments}
We evaluate the approach for the following questions-
\begin{itemize}
        \item Does additional information regarding Level-2 map $\mathbf{F_{t}}$ and subgoal $\mathbf{G^{(2)}_{t}}$ improve safety in complex environments? 
        \item Does \ourmethod{} reliably predict Level-2 map $\mathbf{F_{t}}$ and goal $\mathbf{G^{(2)}_{t}}$?
        \item Does the predicted Level-2 map $\mathbf{F_{t}}$ and goal $\mathbf{G^{(2)}_{t}}$ improve navigation performance in terms of safety compared to using Level-1 $\mathbf{M_{t}}$?
\end{itemize}
\subsection{Experimental Setup}
\textbf{Local Planner and Cost Function Parameters:} We use Log-MPPI~\cite{mohamed2022autonomous} as the local planner for all experiments. The control-noise distribution has standard deviations $\Sigma$ = [0.5, 0.1] for linear and angular velocities, the time horizon was selected to be 10 seconds with the time discretization of 0.1, and the temperature parameter is set to $\lambda$ = 0.5. We use the same cost function as in \cite{poyrazoglu2025unsupervised}. The cost weights are set to $\lambda_{\text{goal}}$ = 200 and $\lambda_{\text{obs}}$ = 50. We modify the distance-to-goal cost $\mathcal{C}_{\text{goal}}$ to incorporate distances to subgoals of both level-1 $\mathbf{G^{(1)}_{t}}$ and level-2 $\mathbf{G^{(2)}_{t}}$.
\par \textbf{Simulation Setup and Metrics}:We evaluate our end-to-end navigation experiments on the test split of the Visdiff~\cite{moorthy2025visdiffsdfguidedpolygongeneration} polygon dataset. We select the start and goal based on the two farthest points with respect to link distance to simulate maximum visibility changes during navigation. We report success rate as a function of link diameter, which defines visibility complexity. Success rate is selected as our safety metric as it measures collision-free navigation from start to goal. All the experiments are performed with maximum velocities of $2 \mathrm{m/s}$ and $3 \mathrm{m/s}$ 
using a differential drive robot. 
\par \textbf{Assumption}:In all the navigation experiments, we assume access to an oracle with full knowledge of the environment, goal, and robot state. At each time step $t$, the oracle provides instructions $\mathbf{I_t}$ and the Level-1 subgoal $\mathbf{G^{(1)}_t}$ in the robot frame. 

\subsection{Ground Truth Performance}
\begin{figure}[h!]
    \centering
    \includegraphics[trim=3.5cm 3.5cm 2.5cm 2.5cm, clip, width=\linewidth]{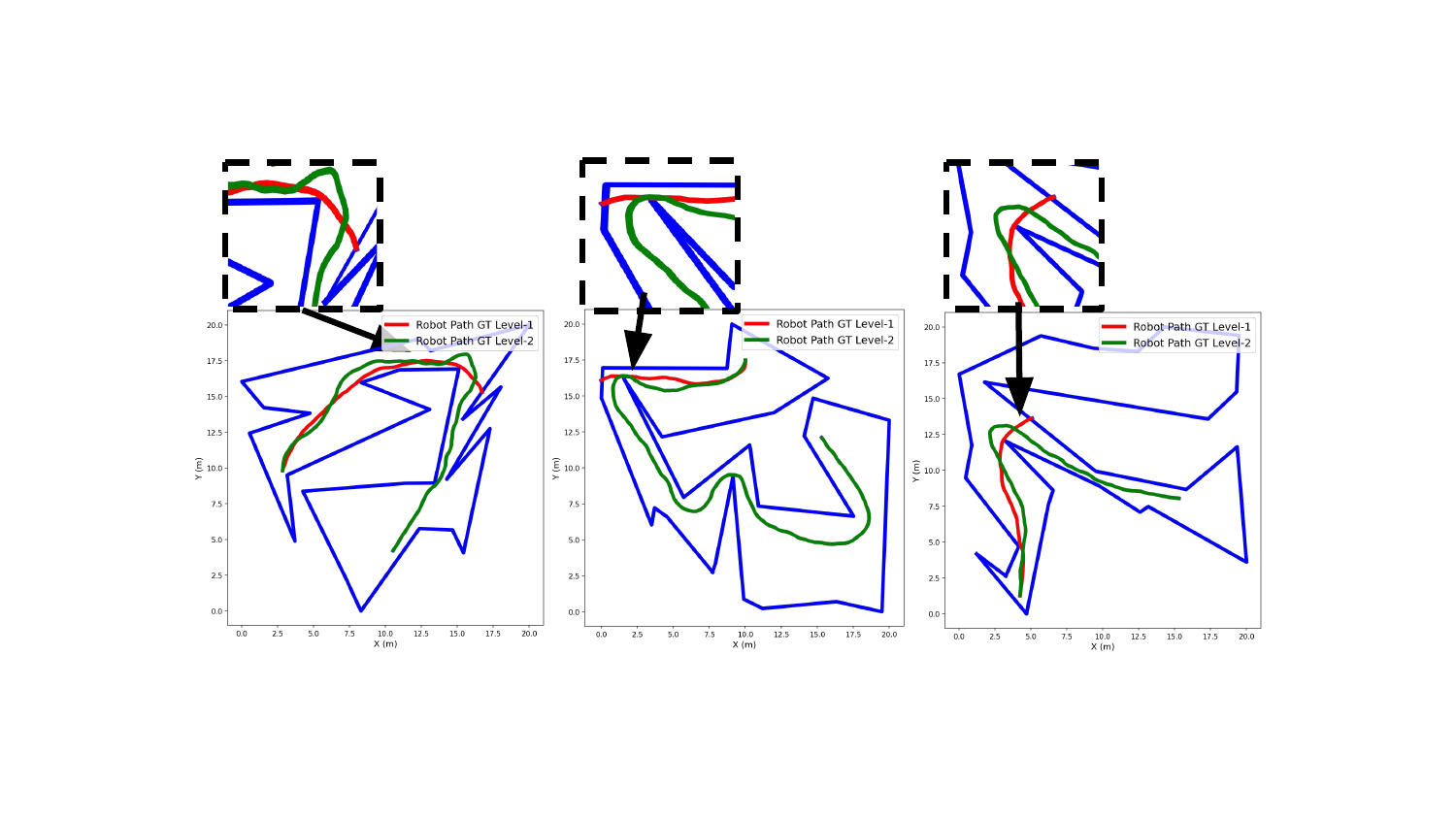} 
    \caption{GT Level-2 vs Local Map Navigation Qualitative Results: At high speed ($3\,\mathrm{m/s}$) in narrow passages, access to ground-truth Level-2 (GT-L2) enables planning ahead and successful turning. Zoomed-in views show that the GT-L2 trajectory (green) approaches the corner with an appropriate entry angle, whereas relying only on the local map (red) fails to complete the turn.
    }
    \label{fig:Qualitative_GT}
\end{figure}

We investigate whether access to the ground truth Level-2 map $\mathbf{F_{t}}$ and subgoal $\mathbf{G^{(2)}_{t}}$ improves the performance of local planners in complex environments. We evaluate by performing end-to-end navigation experiments assuming access to the ground truth $\mathbf{F_{t}}$ and $\mathbf{G^{(2)}_{t}}$. Table~\ref{tab:success_rate}(a) shows the performance of local planners with access to Level-1 $\mathbf{M_{t}}$ and $\mathbf{G^{(1)}_{t}}$, while (b) shows the performance with the additional knowledge of ground truth Level-2 map and subgoal. We observe that access to Level-2 increases the average success rate of local planners by \textbf{46\%} over all link diameters when the maximum allowable velocity is 3 m/s. Figure~\ref{fig:Qualitative_GT} shows the scenario of a sharp turn at high velocity where access to only Level-1 fails to complete the turn. In Level-2, since the planner has prior knowledge of the turn. It plans the path to take a wider turn radius to navigate the sharp turn successfully.

\subsection{Prediction Performance}
\begin{figure}[h!]
    \centering
    \includegraphics[ width=\linewidth]{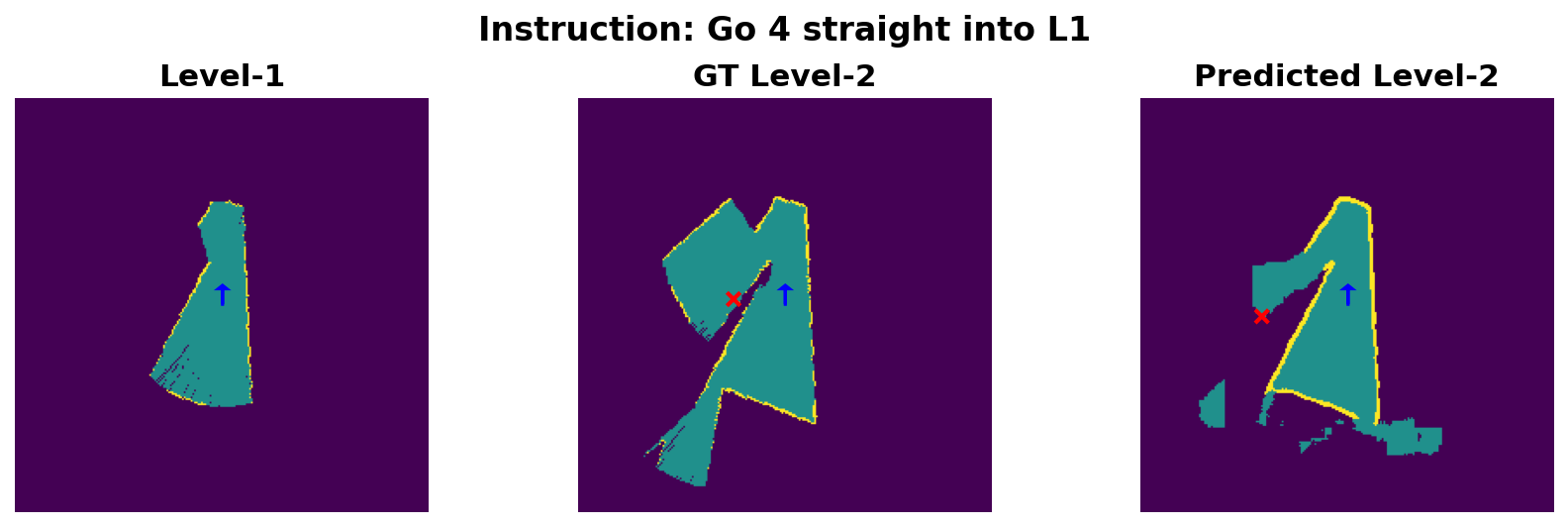} 
    \caption{Unimodal Prediction Qualitative: This result shows the model’s prediction of the Level-2 environment and goal, given that only one opening is visible in the environment and the language instruction specifies going straight into the L1 turn. The blue arrow shows the robot, while the red cross marks the goals. Violet refers to the unknown area in the occupancy grid, yellow refers to the obstacle area, and blue refers to the free space.
    }
    \label{fig:Qualitative_Prediction_uni}
\end{figure}

\begin{figure}[h]
    \centering
    \includegraphics[ width=\linewidth]{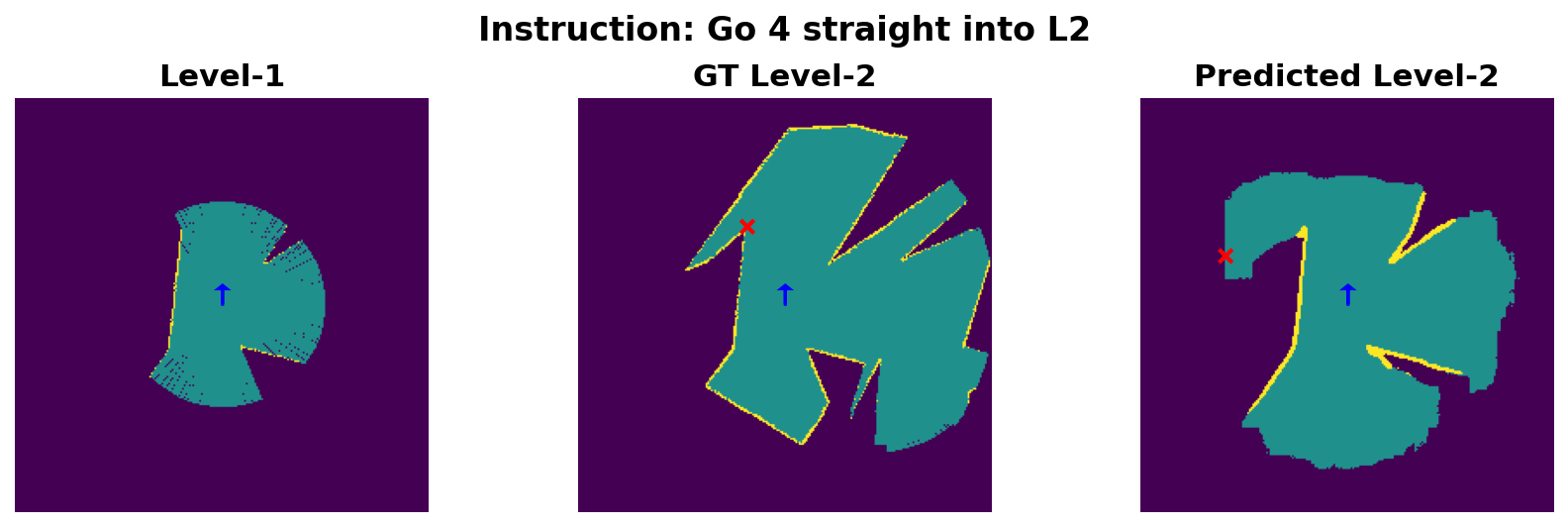} 
    \caption{Multimodal Prediction: This result shows the model’s prediction of the Level-2 environment and goal, given that multiple openings are visible in the environment and the language instruction specifies going straight into the L2 turn. \ourmethod{} generates both the turns and predicts the goal in the proper branch. The blue arrow shows the robot, while the red cross marks the goals. Violet refers to the unknown area in the occupancy grid, yellow refers to the obstacle area, and blue refers to the free space. 
    }
    \label{fig:Qualitative_Prediction_multi}
\end{figure}

We now assess the performance of the predictions generated by \ourmethod{}. We evaluate this with an ablation study over different combinations of Level-2 map and goal sources. In particular, we perform two studies: (i) Level-2 GT map, and predicted subgoal, (ii) Level-2 predicted map and GT subgoal. Table \ref{tab:success_rate} (c) and (d) show the results of these two studies. The Predicted Level-2 map gives an average gain of \textbf{32\%}, while using the predicted subgoal gives a gain of \textbf{40\%} over all link diameters, compared to using only Level-1 map at a velocity of 3 m/s. These gains indicates that the predictions enable local planners navigate reliably in complex environments. Figures~\ref{fig:Qualitative_Prediction_uni} and \ref{fig:Qualitative_Prediction_multi} present qualitative results of \ourmethod{}’s predictions when single and multiple turns are visible. The figures show that \ourmethod{} accurately predicts turns and selects the appropriate branch for successful navigation.

\begin{table*}[h!]
\centering
\resizebox{\textwidth}{!}{%
\begin{tabular}{|c|ccc|cc|cc|cc|cc|}
\hline
\multirow{3}{*}{Case} & \multicolumn{3}{c|}{Method} & \multicolumn{8}{c|}{Success Rate} \\
\cline{2-4}\cline{5-12}
 & L1 & L2 Map & L2 Subgoal & \multicolumn{2}{c|}{Link Diameter 4} & \multicolumn{2}{c|}{Link Diameter 5} & \multicolumn{2}{c|}{Link Diameter 6} & \multicolumn{2}{c|}{Link Diameter 7} \\
\cline{5-12}
 & & & & 2 m/s & 3 m/s & 2 m/s & 3 m/s & 2 m/s & 3 m/s & 2 m/s & 3 m/s \\
\hline
(a) & GT   & --   & --   & 0.92 & 0.84 & 0.96 & 0.86 & 0.84 & 0.52 & 0.80 & 0.42 \\ \hline
(b) & GT   & GT   & GT   & \textbf{0.96} & \textbf{0.94} & \textbf{1.0} & \textbf{1.0} & \textbf{0.92} & \textbf{0.9} & \textbf{0.88} & \textbf{0.88} \\ \hline
(c) & GT   & Pred & GT   & 0.96 & 0.96 & 1.0 & 1.0 & 0.86 & 0.9 & 0.82 & 0.74 \\ \hline
(d) & GT   & GT   & Pred & 0.94 & 0.94 & 1.0 & 0.9 & 0.96 & 0.9 & 0.92 & 0.82 \\ \hline
(e) & GT   & Pred & Pred & 0.92 & 0.94 & 0.98 & 1.0 & 0.84 & 0.9 & 0.82 & 0.78 \\ \hline
\end{tabular}
}
\caption{Success rate by link diameter and maximum velocity. Velocity is shown as subcolumns under each link diameter. GT: ground truth, Pred: predicted, “--” indicates not used, L1: Level - 1, L2: Level-2}
\label{tab:success_rate}
\end{table*}

\subsection{End-to-End Navigation Experiment}
\begin{figure}[h]
    \centering
    \includegraphics[trim=3.5cm 3.5cm 2.5cm 2.5cm, clip, width=\linewidth]{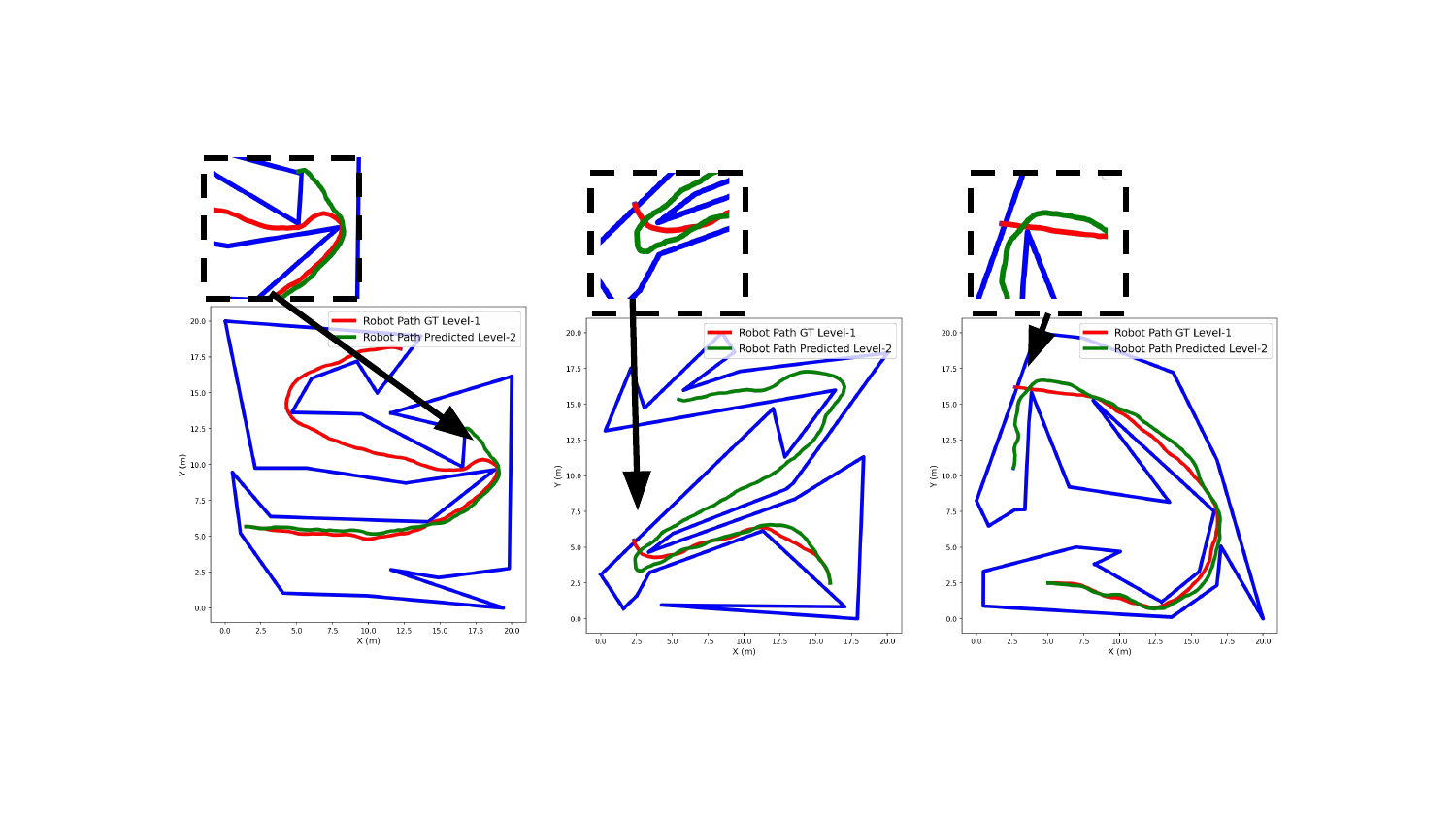} 
    \caption{Predicted Level-2 vs Local Map Navigation Qualitative Results: At high speed (3 m/s) in narrow passages, access
to predicted Level-2 (Pred-L2) enables planning ahead and
successful turning. Zoomed-in views show that the Pred-L2
trajectory (green) approaches the corner with an appropriate
entry angle, whereas relying only on the local map (red) fails
to complete the turn
    }
    \label{fig:Qualitative_Pred}
\end{figure}

We perform end-to-end navigation experiments on the Visdiff dataset using both the predicted Level-2 map and subgoal. Table~\ref{tab:success_rate} (e) shows the results of using both the predicted Level-2 map and subgoal. We observe that with the predictions, we achieve an average gain of \textbf{36\%} over all link diameters. Figure \ref{fig:Qualitative_Pred} shows the scenario of a sharp turn at high velocity where access to only Level-1 fails to complete the turn, while with predicted Level-2, prior knowledge of the upcoming turn allows the planner to plan the path to take a wider turn radius to navigate the sharp turn successfully.

\subsection{Gazebo Simulation Experiments}
\begin{figure}[h!]
  \centering
  \begin{subfigure}{0.48\linewidth}
    \centering
    \includegraphics[width=\linewidth]{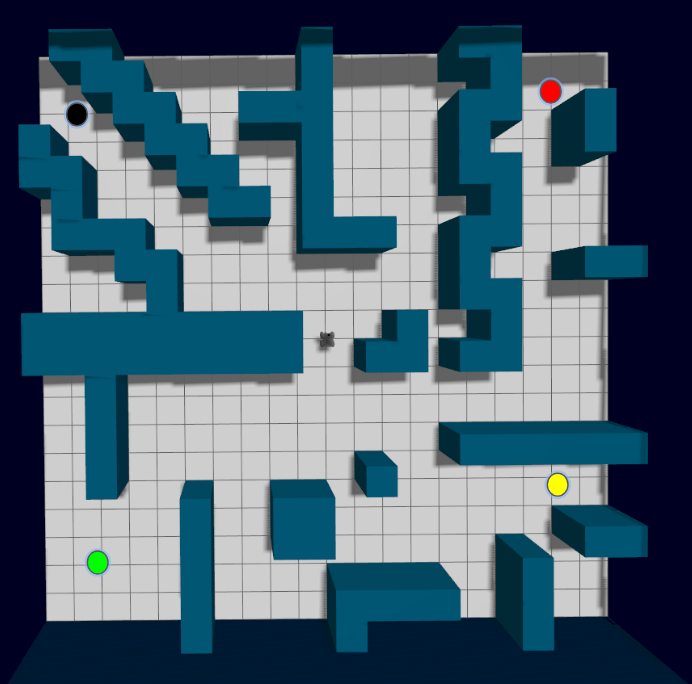}
    \caption{Gazebo Simulation Experiment Setup}
    \label{fig:left}
  \end{subfigure}\hfill
  \begin{subfigure}{0.48\linewidth}
    \centering
    \includegraphics[width=\linewidth]{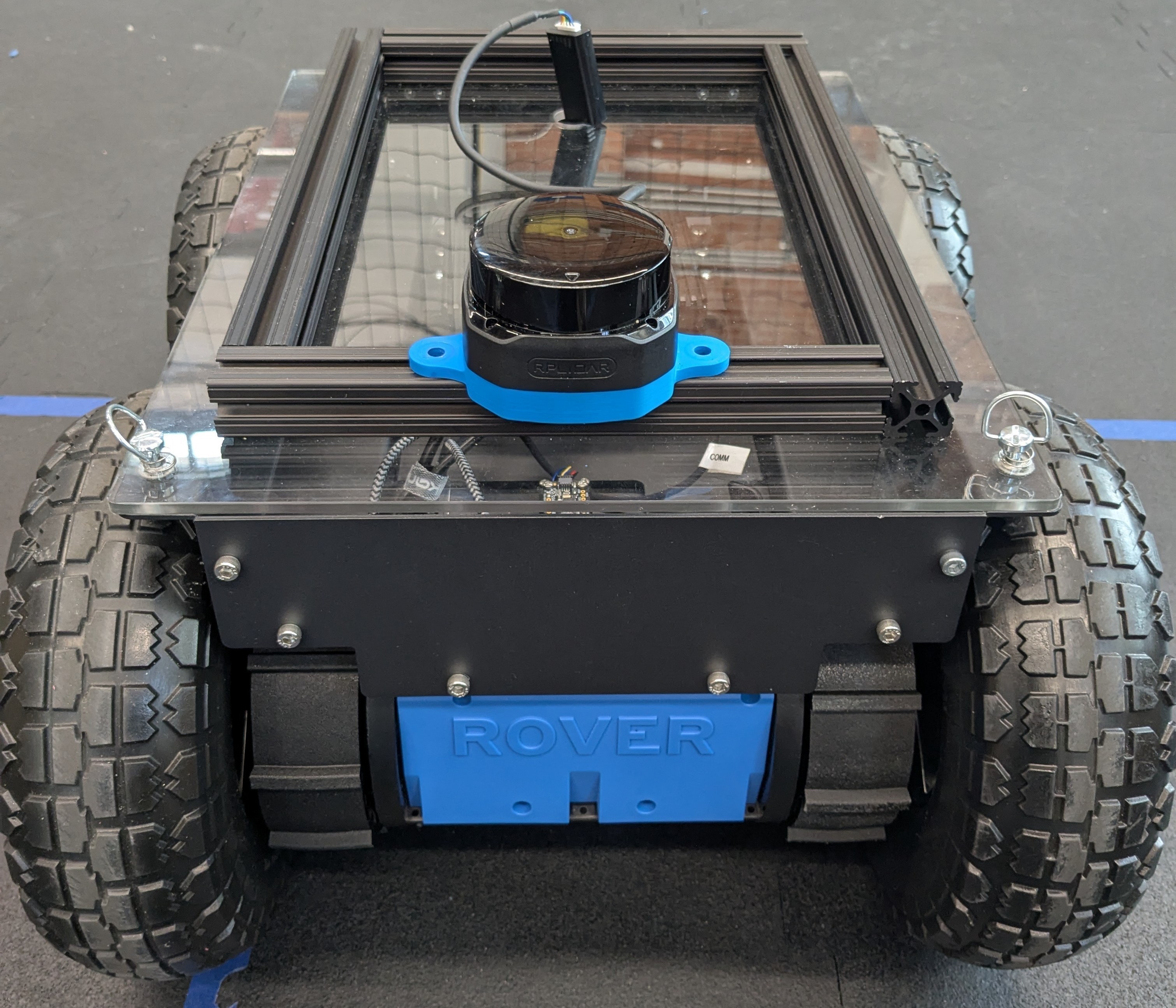}
    \caption{Rover Robot Real Hardware Setup}
    \label{fig:right}
  \end{subfigure}
  \caption{Gazebo Simulation Setup: Figure~(a) shows the maze environment in which gazebo simulation experiments were performed. The four circles in red, yellow, green and  black represnet the goal locations which the robot has to reach. Figure~(b) shows the  hardware configuration of our real-robot which is spawned in the gazebo simulation}
  \label{fig:gazebo_sim}
\end{figure}

We perform Gazebo simulation experiments using the same robot configuration as our real hardware. We evaluate performance in a maze environment chosen for its highly occluded structure. Figure~\ref{fig:gazebo_sim} shows the simulated maze world with selected goal locations and the physical robot platform, which includes a LiDAR, IMU, wheel odometry, and an on-board Jetson Orin Nano. We perform five trials per goal location and report the mean success rate of \ourmethod{} versus a baseline that uses only Level-1 map. Table \ref{tab:simulation_results} shows the comparison of the two approaches across different goal locations. Our results show that \ourmethod{} yields an average gain of \textbf{20\%} compared to the Level-1 baseline. Videos of our real-world experiments are provided in the supplementary material.


\begin{table}[h!]
\centering
\begin{tabular}{|c|c|c|}
\hline
\multirow{2}{*}{Goals} & \multicolumn{2}{c|}{Comparison} \\ \cline{2-3}
                       & Level-1 & \ourmethod{} \\ \hline
Red    & 20\% (1/5) & 20\% (1/5) \\ \hline
Black  & 60\% (3/5) & \textbf{80\%} (4/5) \\ \hline
Green  & 0\% (0/5) & \textbf{40\%} (2/5) \\ \hline
Yellow & 0\% (0/5) & \textbf{20\%} (1/5) \\ \hline
\end{tabular}
\caption{Success Rate Across Different Goals averaged across 5 trials}
\label{tab:simulation_results}
\end{table}

\subsection{Natural Language to Symbolic Instruction Conversion}
\begin{figure}[h!]
    \centering
    \includegraphics[width=\linewidth]{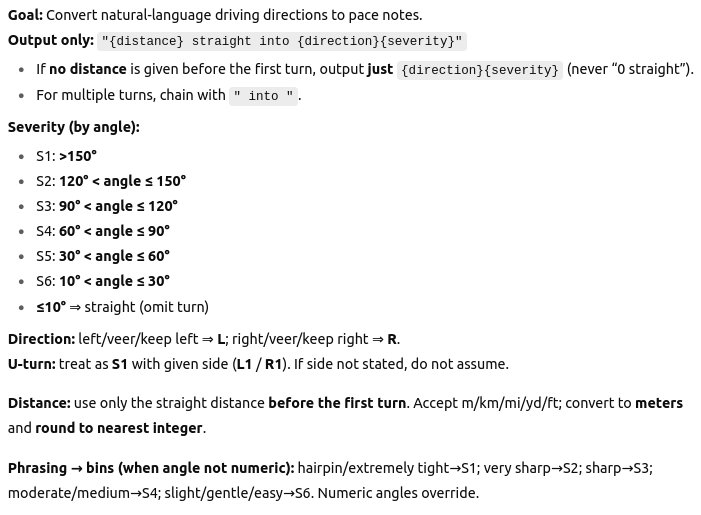} 
    \caption{Natural Language to Symbolic Instruction Prompt: The finalized prompt from ChatGPT-assisted iteration.}
    \label{fig:chatgpt_prompt}
\end{figure}

\begin{figure}[h!]
    \centering
    \includegraphics[width=\linewidth]{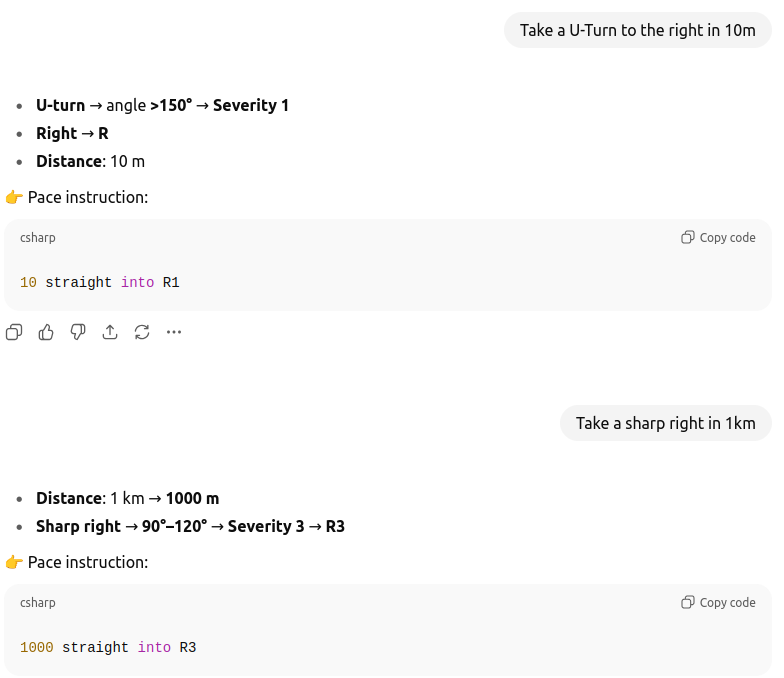} 
    \caption{Prompt Output Samples: The sample output in symbolic language generated by the prompt given the natural language.}
    \label{fig:samples_prompt}
\end{figure}

In real-world applications, conveying symbolic language directly to a robot is challenging without prior training. Therefore, in this section we present an approach that leverages a Large Language Model (LLM) to generate symbolic instructions from natural language. In particular, we use ChatGPT to create the prompt. Figure \ref{fig:chatgpt_prompt} and \ref{fig:samples_prompt} show the prompt and some examples of the natural language to our symbolic representation conversion.

\section{Conclusion}
We presented \ourmethod{}, a novel navigation approach that integrates high level instructions with local planners to address the safety–speed trade-off in cluttered, partially observed environments. We showed that by forecasting Level-2 maps beyond the immediate sensor horizon and predicting instruction-conditioned subgoals enabled planners to act more decisively while remaining safe. Our integration with a Log-MPPI controller showed that \ourmethod{} yields a 36\% improvement over a local-map-only baseline in complex polygonal environments. Lastly, we also presented an approach to convert natural language instructions into our symbolic representation used for language conditioning.
\par In future work, we plan to investigate suitable representations for encoding terrain information conveyed in co-pilot instructions which will enable planners to reason jointly about map and surface conditions.


\addtolength{\textheight}{-12cm}   




\bibliographystyle{unsrt}
\bibliography{root}

@misc{ho2025mapexindoorstructureexploration,
      title={MapEx: Indoor Structure Exploration with Probabilistic Information Gain from Global Map Predictions}, 
      author={Cherie Ho and Seungchan Kim and Brady Moon and Aditya Parandekar and Narek Harutyunyan and Chen Wang and Katia Sycara and Graeme Best and Sebastian Scherer},
      year={2025},
      eprint={2409.15590},
      archivePrefix={arXiv},
      primaryClass={cs.RO},
      url={https://arxiv.org/abs/2409.15590}, 
}

@misc{gao2025mappingsenselightweightneural,
      title={Mapping at First Sense: A Lightweight Neural Network-Based Indoor Structures Prediction Method for Robot Autonomous Exploration}, 
      author={Haojia Gao and Haohua Que and Kunrong Li and Weihao Shan and Mingkai Liu and Rong Zhao and Lei Mu and Xinghua Yang and Qi Wei and Fei Qiao},
      year={2025},
      eprint={2504.04061},
      archivePrefix={arXiv},
      primaryClass={cs.RO},
      url={https://arxiv.org/abs/2504.04061}, 
}

@misc{katyal2020highspeedrobotnavigationusing,
      title={High-Speed Robot Navigation using Predicted Occupancy Maps}, 
      author={Kapil D. Katyal and Adam Polevoy and Joseph Moore and Craig Knuth and Katie M. Popek},
      year={2020},
      eprint={2012.12142},
      archivePrefix={arXiv},
      primaryClass={cs.RO},
      url={https://arxiv.org/abs/2012.12142}, 
}

@inproceedings{wei2021occupancy,
  title={Occupancy map inpainting for online robot navigation},
  author={Wei, Minghan and Lee, Daewon and Isler, Volkan and Lee, Daniel},
  booktitle={2021 IEEE International Conference on Robotics and Automation (ICRA)},
  pages={8551--8557},
  year={2021},
  organization={IEEE}
}

@inproceedings{wang2024agrnav,
  title={Agrnav: Efficient and energy-saving autonomous navigation for air-ground robots in occlusion-prone environments},
  author={Wang, Junming and Sun, Zekai and Guan, Xiuxian and Shen, Tianxiang and Zhang, Zongyuan and Duan, Tianyang and Huang, Dong and Zhao, Shixiong and Cui, Heming},
  booktitle={2024 IEEE International Conference on Robotics and Automation (ICRA)},
  pages={11133--11139},
  year={2024},
  organization={IEEE}
}

@article{huang2022visual,
  title={Visual language maps for robot navigation},
  author={Huang, Chenguang and Mees, Oier and Zeng, Andy and Burgard, Wolfram},
  journal={arXiv preprint arXiv:2210.05714},
  year={2022}
}

@article{goetting2024end,
  title={End-to-end navigation with vision language models: Transforming spatial reasoning into question-answering},
  author={Goetting, Dylan and Singh, Himanshu Gaurav and Loquercio, Antonio},
  journal={arXiv preprint arXiv:2411.05755},
  year={2024}
}

@article{du2025vl,
  title={VL-Nav: Real-time Vision-Language Navigation with Spatial Reasoning},
  author={Du, Yi and Fu, Taimeng and Chen, Zhuoqun and Li, Bowen and Su, Shaoshu and Zhao, Zhipeng and Wang, Chen},
  journal={arXiv preprint arXiv:2502.00931},
  year={2025}
}

@inproceedings{shah2023lm,
  title={Lm-nav: Robotic navigation with large pre-trained models of language, vision, and action},
  author={Shah, Dhruv and Osi{\'n}ski, B{\l}a{\.z}ej and Levine, Sergey and others},
  booktitle={Conference on robot learning},
  pages={492--504},
  year={2023},
  organization={PMLR}
}

@article{ho2022classifier,
  title={Classifier-free diffusion guidance},
  author={Ho, Jonathan and Salimans, Tim},
  journal={arXiv preprint arXiv:2207.12598},
  year={2022}
}

@article{mohamed2022autonomous,
  title={Autonomous navigation of agvs in unknown cluttered environments: log-mppi control strategy},
  author={Mohamed, Ihab S and Yin, Kai and Liu, Lantao},
  journal={IEEE Robotics and Automation Letters},
  volume={7},
  number={4},
  pages={10240--10247},
  year={2022},
  publisher={IEEE}
}

@article{poyrazoglu2025unsupervised,
  title={An Unsupervised C-Uniform Trajectory Sampler with Applications to Model Predictive Path Integral Control},
  author={Poyrazoglu, O Goktug and Moorthy, Rahul and Cao, Yukang and Chastek, William and Isler, Volkan},
  journal={arXiv preprint arXiv:2503.05819},
  year={2025}
}

@misc{moorthy2025visdiffsdfguidedpolygongeneration,
      title={VisDiff: SDF-Guided Polygon Generation for Visibility Reconstruction and Recognition}, 
      author={Rahul Moorthy and Jun-Jee Chao and Volkan Isler},
      year={2025},
      eprint={2410.05530},
      archivePrefix={arXiv},
      primaryClass={cs.CG},
      url={https://arxiv.org/abs/2410.05530}, 
}

@inproceedings{perille2020benchmarking,
title = {Benchmarking Metric Ground Navigation},
author = {Perille, Daniel and Truong, Abigail and Xiao, Xuesu and Stone, Peter},
booktitle = {2020 IEEE International Symposium on Safety, Security and Rescue Robotics (SSRR)},
year = {2020},
organization = {IEEE}
}

@inproceedings{nair2022dynabarn,
  title={Dynabarn: Benchmarking metric ground navigation in dynamic environments},
  author={Nair, Anirudh and Jiang, Fulin and Hou, Kang and Xu, Zifan and Li, Shuozhe and Xiao, Xuesu and Stone, Peter},
  booktitle={2022 IEEE International Symposium on Safety, Security, and Rescue Robotics (SSRR)},
  pages={347--352},
  year={2022},
  organization={IEEE}
}

@inproceedings{cruz2021zillow,
  title={Zillow indoor dataset: Annotated floor plans with 360deg panoramas and 3d room layouts},
  author={Cruz, Steve and Hutchcroft, Will and Li, Yuguang and Khosravan, Naji and Boyadzhiev, Ivaylo and Kang, Sing Bing},
  booktitle={Proceedings of the IEEE/CVF conference on computer vision and pattern recognition},
  pages={2133--2143},
  year={2021}
}

@article{survey,
author = {Tan, Aaron Hao and Nejat, Goldie},
title = {Enhancing Robot Task Completion Through Environment and Task Inference: A Survey from the Mobile Robot Perspective},
year = {2022},
issue_date = {Dec 2022},
publisher = {Kluwer Academic Publishers},
address = {USA},
volume = {106},
number = {4},
issn = {0921-0296},
url = {https://doi.org/10.1007/s10846-022-01776-0},
doi = {10.1007/s10846-022-01776-0},
journal = {J. Intell. Robotics Syst.},
month = dec,
numpages = {24}
}

@article{luperto2023mapping,
  title={Mapping beyond what you can see: Predicting the layout of rooms behind closed doors},
  author={Luperto, Matteo and Amadelli, Federico and Di Berardino, Moreno and Amigoni, Francesco},
  journal={Robotics and Autonomous Systems},
  volume={159},
  pages={104282},
  year={2023},
  publisher={Elsevier}
}

@misc{huang2023audiovisuallanguagemaps,
      title={Audio Visual Language Maps for Robot Navigation}, 
      author={Chenguang Huang and Oier Mees and Andy Zeng and Wolfram Burgard},
      year={2023},
      eprint={2303.07522},
      archivePrefix={arXiv},
      primaryClass={cs.RO},
      url={https://arxiv.org/abs/2303.07522}, 
}

@inproceedings{ratliff2009chomp,
  title={CHOMP: Gradient optimization techniques for efficient motion planning},
  author={Ratliff, Nathan and Zucker, Matt and Bagnell, J Andrew and Srinivasa, Siddhartha},
  booktitle={2009 IEEE international conference on robotics and automation},
  pages={489--494},
  year={2009},
  organization={IEEE}
}

@article{lavalle2001randomized,
  title={Randomized kinodynamic planning},
  author={LaValle, Steven M and Kuffner Jr, James J},
  journal={The international journal of robotics research},
  volume={20},
  number={5},
  pages={378--400},
  year={2001},
  publisher={SAGE Publications}
}

@article{labbe2019rtab,
  title={RTAB-Map as an open-source lidar and visual simultaneous localization and mapping library for large-scale and long-term online operation},
  author={Labb{\'e}, Mathieu and Michaud, Fran{\c{c}}ois},
  journal={Journal of field robotics},
  volume={36},
  number={2},
  pages={416--446},
  year={2019},
  publisher={Wiley Online Library}
}

@misc{tao2023learningexploreindoorenvironments,
      title={Learning to Explore Indoor Environments using Autonomous Micro Aerial Vehicles}, 
      author={Yuezhan Tao and Eran Iceland and Beiming Li and Elchanan Zwecher and Uri Heinemann and Avraham Cohen and Amir Avni and Oren Gal and Ariel Barel and Vijay Kumar},
      year={2023},
      eprint={2309.06986},
      archivePrefix={arXiv},
      primaryClass={cs.RO},
      url={https://arxiv.org/abs/2309.06986}, 
}

@misc{zwecher2022integratingdeepreinforcementsupervised,
      title={Integrating Deep Reinforcement and Supervised Learning to Expedite Indoor Mapping}, 
      author={Elchanan Zwecher and Eran Iceland and Sean R. Levy and Shmuel Y. Hayoun and Oren Gal and Ariel Barel},
      year={2022},
      eprint={2109.08490},
      archivePrefix={arXiv},
      primaryClass={cs.LG},
      url={https://arxiv.org/abs/2109.08490}, 
}

@misc{sharma2023proxmapproximaloccupancymap,
      title={ProxMaP: Proximal Occupancy Map Prediction for Efficient Indoor Robot Navigation}, 
      author={Vishnu Dutt Sharma and Jingxi Chen and Pratap Tokekar},
      year={2023},
      eprint={2203.04177},
      archivePrefix={arXiv},
      primaryClass={cs.RO},
      url={https://arxiv.org/abs/2203.04177}, 
}

@misc{chen2020topologicalplanningtransformersvisionandlanguage,
      title={Topological Planning with Transformers for Vision-and-Language Navigation}, 
      author={Kevin Chen and Junshen K. Chen and Jo Chuang and Marynel Vázquez and Silvio Savarese},
      year={2020},
      eprint={2012.05292},
      archivePrefix={arXiv},
      primaryClass={cs.RO},
      url={https://arxiv.org/abs/2012.05292}, 
}

@misc{perez2017filmvisualreasoninggeneral,
      title={FiLM: Visual Reasoning with a General Conditioning Layer}, 
      author={Ethan Perez and Florian Strub and Harm de Vries and Vincent Dumoulin and Aaron Courville},
      year={2017},
      eprint={1709.07871},
      archivePrefix={arXiv},
      primaryClass={cs.CV},
      url={https://arxiv.org/abs/1709.07871}, 
}

\end{document}